\documentclass[10pt,twocolumn,letterpaper]{article}

\usepackage[pagenumbers]{cvpr} % To force page numbers, e.g. for an arXiv version

\usepackage{array}
\usepackage{amsmath}
\usepackage{amssymb}
\usepackage{latexsym}
\usepackage{textcomp}
\usepackage{multicol}
\usepackage{multirow}
\usepackage{arydshln}
\usepackage{helvet}
\usepackage{courier}
\usepackage{amsfonts}
\usepackage{bbm}
\usepackage{diagbox}
\usepackage{booktabs}
\usepackage{bbding}
\usepackage[colorlinks,
            linkcolor=blue,
            anchorcolor=blue,
            citecolor=blue
            ]{hyperref}
\usepackage{cleveref}
\usepackage{graphicx}
\usepackage{bm}
\usepackage{threeparttable}
\usepackage{mathrsfs}
\usepackage{amssymb}
\usepackage{rotating}

\title{Semi-Supervised Video Inpainting with Cycle Consistency Constraints}
\begin{document}

\author{Zhiliang Wu\textsuperscript{\rm 1}, 
Hanyu Xuan\textsuperscript{\rm 1},
Changchang Sun\textsuperscript{\rm 2},
Kang Zhang\textsuperscript{\rm 1},
Yan Yan\textsuperscript{\rm 2}\\
\textsuperscript{\rm 1}
School of Computer Science and Engineering, Nanjing University of Science and Technology, China\\
\textsuperscript{\rm 2}
Department of Computer Science, Illinois Institute of Technology, USA}
% Institution1\\
% Institution1 address\\

\maketitle

\begin{abstract}
% Video inpainting aims to fill corrupted regions of each frame with plausible content filtered from the video sequence. 
%based 或者是related work
%Deep neural networks
Deep learning-based video inpainting has yielded promising results and gained increasing attention from researchers. 
Generally, these methods assume that the corrupted region masks of each frame are known and easily obtained.
However, the annotation of these masks are labor-intensive and expensive,
%mask不能容易得到？ 还是难以得到？
%which is labor-intensive and expensive for real-world applications.
which limits the practical application of current methods.
% However, due to the fact that the annotation of these masks are labor-intensive and expensive, the practical application of current methods are limited.
% because obtaining these masks is labor-intensive and expensive.
%limiting its practical application.
%Recent deep learning-based algorithms have achieved promising results in this field. However, these methods usually assume that the damaged region masks for each frame are known, which limits its practical application because obtaining these masks is labor-intensive and expensive.
%引出你的方法与现有方法的区别 only one frame
%如果能够使用有限个的mask，同时能够保证很好的性能，这样对于现实应用会带来很大的帮助。
% In this paper,
Therefore, 
we expect to
relax this assumption by defining a new semi-supervised inpainting setting, making the networks have the ability of
completing the corrupted regions of the whole video 
using the annotated mask of only one frame. 
Specifically, 
in this work,
we propose an end-to-end trainable framework consisting of 
% which includes a pair of dual networks: 
completion network and mask prediction network,
which are designed to generate corrupted contents of the current frame using the known mask and decide the regions to be filled of the next frame, respectively.
% The former aims to complete corrupted regions indicated by the mask, while the latter generates the mask used to indicate the corrupted regions for unannotated frames. 
% Particularly, 
Besides, we introduce a cycle consistency loss
to regularize the training parameters of these two networks.
In this way, the completion network and the mask prediction network can constrain each other, and hence the overall performance of the trained model can be maximized.
%XXX%携手前进% accurate correspondence
%can be maxmized.
% to capture the accurate correspondence between the completion network and mask prediction network, 
% a cycle consistency loss is introduced c
Furthermore, due to the natural existence
% introduction 
of prior knowledge (e.g., corrupted contents and clear borders), 
current 
% fully-supervised
video inpainting datasets are not suitable
in the context of 
% for 
% the 
semi-supervised video inpainting.
Thus, we create 
a new dataset by simulating the corrupted video of real-world scenarios.
% for the semi-supervised video inpainting task.
 Extensive experimental results are 
 % provided
 reported
 to demonstrate the superiority of our model in the video inpainting task. 
 Remarkably, although our model 
 is trained 
 in a semi-supervised manner, it
 can achieve comparable performance as fully-supervised methods.
%  As a byproduct, we will release the code and dataset to benefit other researchers.
 % The code and dataset will be released.
 %scc0811: 我觉得现在abstract没啥问题了。
\end{abstract}

\section{Introduction}
% Video inpainting is a promising yet challenging task, which aims to fill corrupted regions of videos with plausible contents. 
Video inpainting aims to fill corrupted regions of the video with plausible contents, which is a promising yet challenging task in computer vision.
% which is a promising yet challenging task.
% As a fundamental task in computer vision, video inpainting
It can benefit a wide range of practical applications, such as scratch restoration~\cite{chang2019free}, undesired object removal~\cite{9010390}, and autonomous driving~\cite{liao2020dvi}, etc. 
% From a methodological perspective,
In essence,
unlike image inpainting which usually learns on the spatial dimension, video inpainting task pays more attention to exploiting the temporal information. 
Naively using image inpainting methods~\cite{yan2019PENnet,978-3-030-58595-2_45,9710199,Kim_2022_CVPR} on individual video frame 
%scc: 这里有没有仅仅把image用到video上的方法呢？
to fill corrupted regions will lose inter-frame motion continuity, resulting in flicker artifacts in the inpainted video. 

\begin{figure}[!t]
\centering%height=3.8cm,width=17.5cm
\includegraphics[scale=0.40]{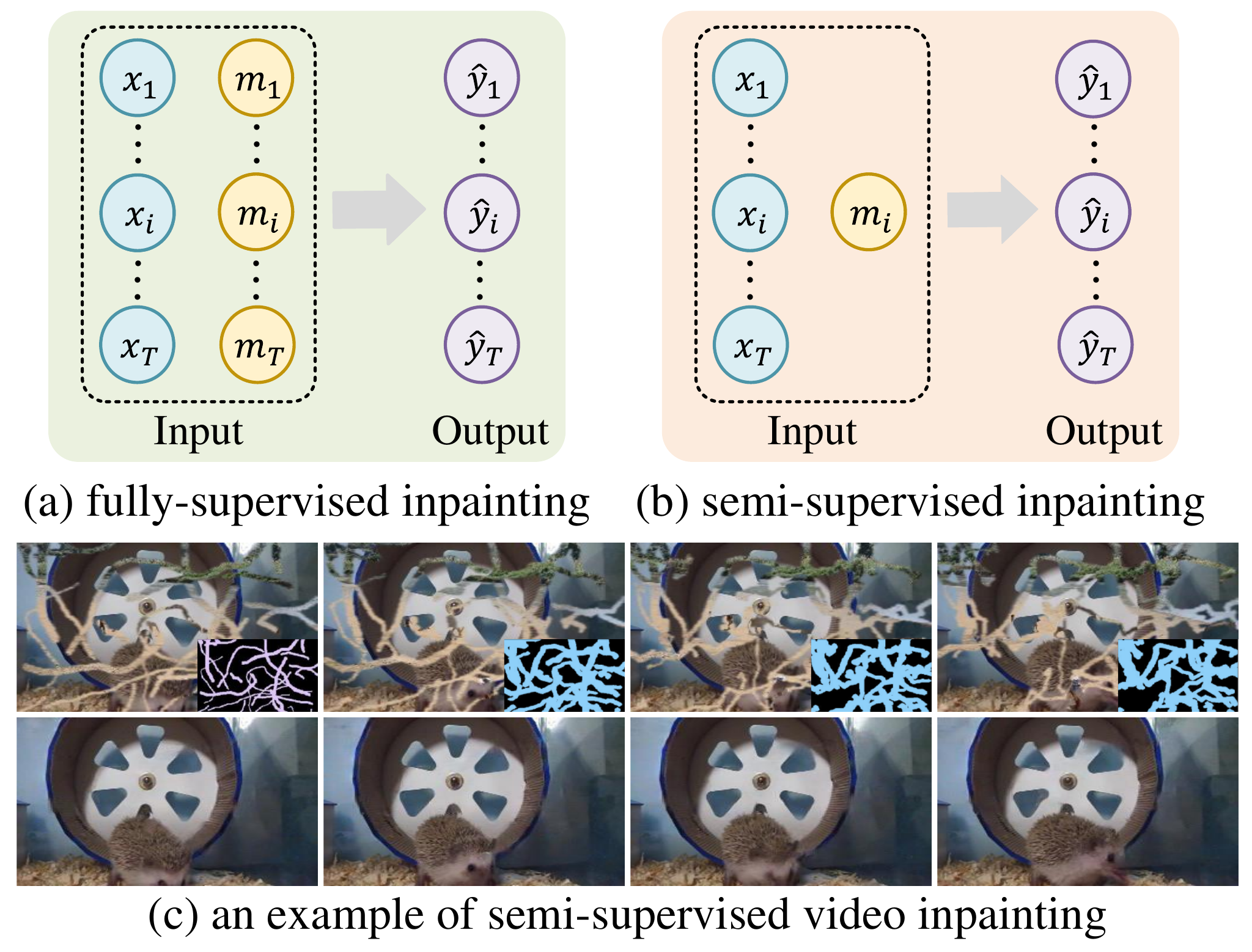}
\vspace{-0.6cm}
\caption{Existing methods perform video inpainting in a fully-supervised setting. The typical issue of such methods is the need to elaborately annotate the corrupted regions $\textbf{\emph{m}}_{i}$ of each frame $\textbf{\emph{x}}_{i}$ in the video (Fig.(a)),  
%scc0812: 文中缺少对（a）和（b）的引用，在适当的位置引用他们可以更好地说明现有方法和你的semi方法之间的差别以及你的好处。画了图就要引用的！！！
which is labor-intensive and expensive 
% for
in real-world applications. In this paper, we formulate a new task: semi-supervised video inpainting, which only annotates the corrupted regions of one frame to complete the whole video (Fig.(b)). 
%scc0812: 说明一下这四张图之间的关系，是前后帧的关系吗？
Fig.(c) shows an example of semi-supervised video inpainting: the top row shows sample frames with the mask, where pink denotes the manually annotated mask of corrupted regions, and blue denotes the mask of corrupted regions predicted by the model. The completed results $\widehat{\textbf{\emph{y}}}_i$ are shown in the bottom row.}
\label{Fig1}
\vspace{-0.5cm}
\end{figure}

%scc:here
Similar to the texture synthesis,
traditional video inpainting methods~\cite{granados2012how,huang2016temporally,newson2014video} attempt to generate missing contents
to fill missing regions
% using ideas similar to texture synthesis, 
% \emph{i.e.}, 
by searching and copying similar patches from known regions. 
Despite the authentic completion results
have been achieved,
these approaches still meet grand challenges, such as the lack of high-level understanding of the video~\cite{wang2018video}
%scc：我有一点疑问 high-level understanding 到底是指哪方面 是不是应该解释的再详细一些。
and high computational complexity~\cite{chang2019free,lee2019cpnet}.
In fact,
recently, deep learning-based video inpainting methods~\cite{chang2019free,Kim_2019_CVPR,lee2019cpnet,wang2018video,xu2019deep,Gao-ECCV-FGVC,zou2021progressive,li2022towards,Ren_2022_CVPR} have 
been noticed by many researchers and 
made promising progress in terms of the 
%scc: 这里的quality应该说明是修复质量或者啥的?
quality and speed. These methods usually
% expect to 
extract relevant features from the neighboring frames by convolutional neural networks and perform different types of context aggregation to generate missing contents. 
% Nevertheless, these methods are only suitable for scenarios where the corrupted region mask for each frame of the video is known.
Nevertheless, these methods are only suitable 
% for 
in the 
scenarios where the corrupted region mask
% for each frame of the video is known.
of each frame in the video is 
% known.
available (Fig.\ref{Fig1}(a)).
%usually 
%适用 However, these methods are only suitable for scenarios where the damaged region mask for each frame of the video is known.
%assume that corrupted region mask of each frame is known, 
Hence, when we resort to these methods in real-world applications,
annotating the corrupted regions of each frame in the video
is required,
which is labor-intensive and expensive, especially for long videos. %scc:你看一下这一句要不要加上
% so we need to elaborately annotate the corrupted regions of each frame in the video when using these models, 
% which is fatal for real-world applications because obtaining these masks is labor-intensive and expensive.
%which is labor-intensive and expensive for real-world applications.
Formally, a naive combination of video object segmentation and video inpainting methods can 
be used to
reduce annotation costs in a two-stage manner.
In this way, we can first 
% masks are first generated 
generate the masks
for all video frames using a video object segmentation method, 
and then complete the missing regions with the fully-supervised video inpainting methods. 
However, such a two-stage approach 
has 
% the following 
some intuitive
disadvantages.
One the one hand,
% the overall performance is sub-optimal since each module is learned individually; 
since each module is learned individually and sometimes not well combined to maximize performance,
%\scc{...},
%scc: 有时候不能地结合实现性能最大化
the overall performance is sub-optimal.
On the other hand,
existing video object segmentation methods are unsatisfactory for segmentation results of the scratch regions similar to Fig.\ref{Fig1}(c), resulting in the inpainted video with critical errors. 
%Therefore, designing a framework that is end-to-end trainable and can handle corrupted regions with complex patterns is necessary for semi-supervised video inpainting tasks.
Therefore, to realize the goal of reducing annotation cost for video inpainting from scratch,
%to realize the task of \scc{xxx} from scratch,
in this paper, we introduce a new semi-supervised video inpainting setting that we can complete the corrupted regions of the whole video using the annotated mask of only one frame (Fig.\ref{Fig1}(b)) and train the network from end-to-end.
%scc0812: 可以看一下section2.1的一个comment
%(\scc{often the first frame of the video}).
%In this paper, we introduce a
%new problem:%%where we can
%semi-supervised video inpainting, which only annotates the corrupted regions of one frame (often the first frame of the video) to complete the whole video. 
In this way,
compared with the conventional fully-supervised setting, 
the annotation cost can be greatly reduced,
making video inpainting more convenient in practical application.
%real-world applications.
%and%增加实际引用的便携和可用}
% a significant advantage of the semi-supervised video inpainting is the great reduction of the annotation cost. 
%scc：对于长视频 更加有意义 more beneficial when it comes to .....
%This beneﬁt becomes more obvious if video inpainting is performed in the long video.
% fenkai

%\wzl{For semi-supervised video inpainting, a naive solution is first to use a video object segmentation method to generate masks for all video frames, and then complete the missing regions with the fully-supervised video inpainting method. However, such a two-stage approach has the following disadvantages: 1) the overall performance is sub-optimal since each module is learned individually; 2) existing video object segmentation methods are unsatisfactory for segmentation results of the scratch regions similar to Fig.\ref{Fig1}(c), usually resulting in the inpainted video with critical errors. Therefore, designing a framework that is end-to-end trainable and can handle corrupted regions with complex patterns is necessary for semi-supervised video inpainting tasks.}

However, fulfilling the task of semi-supervised video inpainting is non-trivial and has some issues to be addressed.
On the one hand,
%scc：我记得好像是except for或者expect that。。 不是except sth
% except for the annotated one frame, 
%scc0812:
except for one annotated known frame, 
there are 
no masks for other frames to indicate the corrupted regions in the proposed semi-supervised setting.
% The difficulty of the proposed semi-supervised video inpainting task is that the video's other frames (except the annotated
% frame) have no mask used to indicate the corrupted regions. 
% To solve the problem, 
To solve this problem, 
we decompose the semi-supervised video inpainting task into 
% a pair of dual tasks: %scc：一对双重任务？
dual tasks:
frame completion and mask prediction. 
% Specifically, we first perform frame completion on the given mask and the corresponding frame to obtain the completed frame and then feed it and the subsequent frame into the mask prediction network to generate the corrupted regions mask of the subsequent frame. 
Specifically, we first perform frame completion 
% on the given mask and the corresponding frame %scc：补全应该是在帧上而不是mask
on the frame with corresponding given mask
to obtain the completed frame using the designed frame completion network.
Then, we 
% and then 
%scc0812: 这里的it换成 the completed frame 会不会更好一些？
feed the completed frame
%\scc{it} 
and the subsequent frame into the 
proposed mask prediction network to generate the corrupted region mask of the subsequent frame.
Last,
by iterating frame by frame, we can complete corrupted regions of each frame in the video. 
On the other hand,
to precisely
capture the accurate correspondence between the completion network and mask prediction network, a cycle consistency loss is introduced to regularize the trained parameters. 
% Furthermore,
In addition, existing video inpainting datasets usually take black or noise pixels as the corrupted contents of the video frame.
% However, 
In fact, %scc：我觉得把下面当做一个事实进行陈述，会更好一些
such a setting will introduce some specific prior knowledge (e.g., corrupted contents and clear borders) into the dataset, 
%making the mask prediction network easily distinguish corrupted regions from natural images. 
making it easy for the mask prediction network to distinguish corrupted regions from natural images.
In this way,
% Therefore, 
existing datasets cannot 
% effectively 
realistically 
simulate complex real-world scenarios.
%scc0812: 我想了想又有点迷惑，这些balck和noise pixels的图片和一般被corrupt的图片之间的区别是什么呢？ 我觉得需要说明一下这些black是人工annoted吗？是为了这个任务人随机污染的图片吗？
Hence, in our work, to effectively avoid the introduction of the above prior knowledge into the dataset,
% In this scenario, 
we use natural images as corrupted contents of the video frame
%scc:我们使用自然图像作为损坏的内容. 这句话不通顺
and apply iterative Gaussian smoothing~\cite{wang2018image} to extend the edges of corrupted regions.
% effectively avoiding the introduction of the above prior knowledge into the dataset. 
Experimental results demonstrate that our proposed method can achieve comparable inpainting results as fully-supervised methods. 
An example result of our method is shown in Fig.\ref{Fig1}(c).
Our contributions are summarized as follows:
\begin{itemize}
  % \item A novel semi-supervised video inpainting task is formulated. It aims to complete the corrupted regions of the whole video given the mask of one frame, which can reduce the annotation cost of video inpainting in practical applications. To the best of our knowledge, this is the first work for semi-supervised video inpainting task.
    \item 
    %To reduce the annotation cost of video inpainting in practical applications,
    We formulate a novel semi-supervised video inpainting task that aims to complete the corrupted regions of the whole video with the given mask of one frame.
    % , which can reduce the annotation cost of video inpainting in practical applications. 
    To the best of our knowledge, this is the first end-to-end semi-supervised work in the video inpainting field.
  \item A flexible and efficient framework consisting of completion network and mask prediction network is designed to solve the semi-supervised video inpainting task,
%   Meanwhile, 
  where cycle consistency loss is introduced to regularize the trained parameters.
  % The framework contains a pair of dual networks: completion network and mask prediction network, and a cycle consistency loss is introduced to regularize the training parameters.
  %\item To capture the accurate correspondence between the completion network and mask prediction network, a cycle consistency loss is introduced to regularize the training parameters.
  %\item To the best of our knowledge, this is the first work to provide a uniﬁed framework for semi-supervised video inpainting task, which provides a benchmark for subsequent research.
  \item A novel synthetic dataset is tailored for the semi-supervised video inpainting task.
   which consists of 4,453 video clips. This dataset will be published to facilitate subsequent research and benefit other researchers.
\end{itemize}
\begin{figure*}[!t]
\centering%height=3.0cm,width=15.5cm
\includegraphics[scale=0.5]{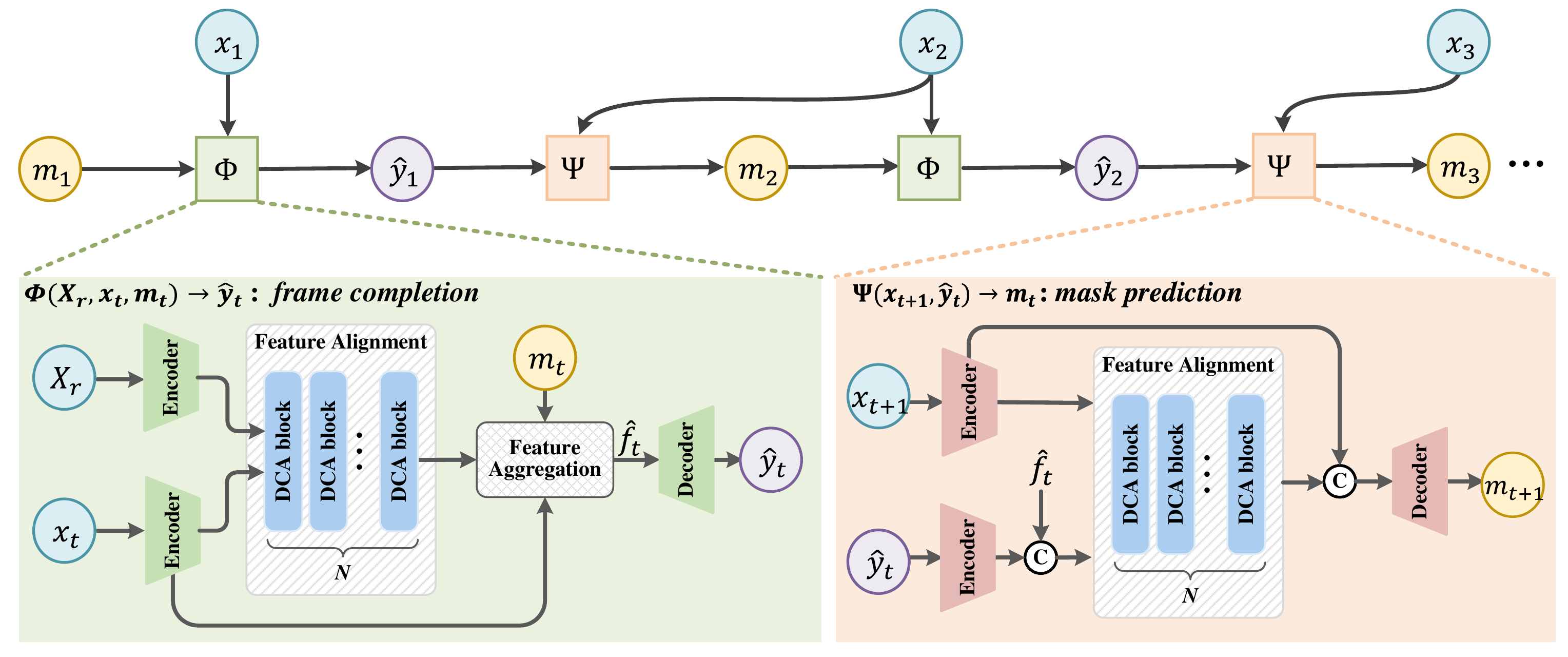}
\vspace{-0.2cm}
\caption{\textbf{Illustration of the proposed semi-supervised video inpainting networks.} Our networks are composed of a $\emph{completion network} (\bm\Phi)$ and a $\emph{mask prediction network} (\bm\Psi)$. The $\emph{completion network}$ uses temporal information of reference frames $\textbf{\emph{X}}_{r}$ to complete the corrupted regions $\textbf{\emph{m}}_{t}$ of the target frame $\textbf{\emph{x}}_{t}$. The $\emph{mask prediction network}$ takes the completed target frame $\widehat{\textbf{\emph{y}}}_{t}$ and frame $\textbf{\emph{x}}_{t+1}$ as input to generate the mask $\textbf{\emph{m}}_{t+1}$, 
% which is used 
which can be adopted
to indicate the corrupted regions of the next frame $\textbf{\emph{x}}_{t+1}$. 
By alternating forward, 
% we can only annotate the mask of one frame to complete the whole video.
%scc0812:
we can complete the whole video with only one annotated frame.
DCA block denotes deformable convolution alignment block. $C$ 
% is
stands for the
concatenate operation.}
\vspace{-0.4cm}
\label{Fig2}
\end{figure*}

\section{Related Works}
%scc0812: 其他section都没有这一段 你看有没有必要保留
% \scc{In this section, we briefly review recent advances in Video Inpainting  and semi-supervised Video Object Segmentation (VOS).} 
\paragraph{Video Inpainting}
can be roughly classified into two lines: patch-based and deep learning-based video inpainting. 

Patch-based inpainting methods~\cite{granados2012how,huang2016temporally,newson2014video} 
% solved
solve the video inpainting task by searching and pasting coherent contents from the known regions into the corrupted regions. However, %these approaches are infeasible
% for 
%in 
% real-time scc0812:你的这个应该不是real-time的任务吧？
%scc08152: 这里为什么要说明real-time呢？ 我看你下面也没有介绍其他可以real-time的相关工作，还是下面介绍的deep-based都是rela-time的？ 你确认一下吧，如果real-world不对
%real-world
%applications because they suffer high computational complexity due to the heavy optimization process~\cite{chang2019free}. 
these approaches often suffer high computational complexity due to the heavy optimization process, which limits their real-world applications~\cite{chang2019free}.

% With the rapid development of deep learning, deep learning-based video inpainting methods have made significant progress. 
Deep learning-based video inpainting mainly focuses on three directions: 1) \emph{3D convolutional networks}~\cite{Kim_2019_CVPR,chang2019free,9558783}, which usually reconstruct the corrupted contents by directly aggregating temporal information from neighbor frames through 3D temporal convolution; 2) \emph{flow guided approaches}~\cite{lao2021flow,xu2019deep,zou2021progressive,Gao-ECCV-FGVC,li2022towards,Zhang_2022_CVPR,Kang2022ErrorCF}, which first use a deep flow completion network to restore the flow sequence, and then use the restored flow sequence to guide the relevant pixels of the neighbor frames into corrupted regions; and 3) \emph{attention-based methods}~\cite{lee2019cpnet,Li2020ShortTermAL,liu2020temporal,liu2021fuseformer,9010390,9446636,Ren_2022_CVPR,zhang2022flow}, which retrieve information from neighbor frames and use weighted sum to generate corrupted contents. 
Although these methods have achieved promising completion results, we still need to elaborately annotate the corrupted regions of each frame in the video, 
% when using them, 
which limits its applications.
%scc0812: 我重新换了一种说法
% Although these methods have achieved promising completion results, 
% the annotation of corrupted regions of each frame in the video is needed, resulting in the limination of its real application. 
Unlike these approaches, 
% the 
our proposed semi-supervised video inpainting only uses the mask of one frame to 
complete the corrupted regions of the whole video.
%scc0812：其实一直有个疑问，你这个方法能不能反向往前预测，如果已知的帧不是第一帧。 我觉得如果可以的话，最好还是说明一下，如果不可以的话，那也要表明一下。
\vspace{-0.5cm}
\paragraph{Semi-supervised Video Object Segmentation}
 is understood as using a single annotated frame (usually the first frame of the sequence) to estimate the object position in the subsequent frames of the video. 
 % The 
 Existing 
 % works of 
 semi-supervised video object segmentation can be broadly classified into three categories, \emph{i.e.}, \emph{matching-based}~\cite{8578250,robinson2020learning,xie2021efficient}, \emph{propagation-based}~\cite{zhang2019fast,huang2020fast,chen2020state}, and \emph{detection-}based methods~\cite{9008825,wang2019fast,sun2020fast}. Matching-based methods~\cite{8578250,robinson2020learning,xie2021efficient} usually
 % trained %scc0812:这种总结性的语句我觉得应该用现在时
 train a typical Siamese matching network to find objects similar to the given mask of the first frame in subsequent frames. 
 Besides, propagation-based methods~\cite{zhang2019fast,huang2020fast,chen2020state}
 % embedded
 work on 
 embedding image pixels into a feature space and utilized temporal information to guide label propagation. 
 In addition,
 Detection-based methods~\cite{9008825,wang2019fast,sun2020fast}
 first
 % learned
 learn feature representation of the annotated objects in the first frame and then
 % detected 
 detect corresponding pixels in subsequent frames. 
 %scc0812: 我重新写了一下这一句，你把内容补全把
In spite of the compelling success achieved by these methods, 
 far too little attention has been paid to the segmentation of scratched regions with complex patterns.
 %\scc{add something}.
 In this situation, we put forward a mask prediction network that utilizes the current completed frame and subsequent frame as 
input to the generate corrupted region mask.
 Fortunately, we find the fact that the only difference between the video frames before and after completion is the corrupted regions of the current frame. Therefore, our mask prediction network is suitable for predicting damaged regions of various modalities, so that making our proposed semi-supervised inpainting framework robust.
 %Therefore, in our work, 
% we put forward a mask prediction network to \scc{add some thing}.
% Our mask prediction network takes the frames before and after the completion as input. ?????scc0812不太懂这是啥意思
%Ideally, the difference between these two frames is only the corrupted regions of the current frame. Therefore, our mask prediction network is suitable for predicting corrupted regions of various patterns.
%scc0812: 上面这一句与前面的承接再看一下
%\scc{add some to introduce your semi video inpainting, like current methods did not explore it in this task.}
\section{Method}
\subsection{Problem Formulation}

%\textbf{\emph{fully-supervised setting vs. semi-supervised setting}}
Let $\textbf{\emph{X}}=\{\textbf{\emph{x}}_1,\textbf{\emph{x}}_2, \dots,\textbf{\emph{x}}_T\}$ be a corrupted video sequence consisting of $T$ frames. $\textbf{\emph{M}}=\{\textbf{\emph{m}}_1,\textbf{\emph{m}}_2, \dots,\textbf{\emph{m}}_T\}$ denotes the corresponding frame-wise masks,
%The corrupted regions in video sequence $\textbf{\emph{X}}$ 
%scc0812: 因为你这篇是semi，所以你的问题定义顺序就不能和另外一篇论文一样了。这里的binary mask是你在已经知道一个mask的情况下迭代学习出来的，所以这个known mask和predicted masks 就应该区别出来，这样才是你semi setting下的问题定义。可能你需要重新组织一下这个部分，注意突出semi
%\scc{are annotated by the binary masks $\textbf{\emph{M}}=\{\textbf{\emph{m}}_1,\textbf{\emph{m}}_2, \dots,\textbf{\emph{m}}_T\}$, 
where the mask $\textbf{\emph{m}}_i$ represents the corrupted regions of frame $\textbf{\emph{x}}_i$.
%For each mask $\textbf{\emph{m}}_i$, 
%%a value of “0” 
%and 
% or 
%“1” indicates the corresponding pixel in the frame $\textbf{\emph{x}}_t$ 
% is
%are
%valid and corrupted (or missing), respectively. 
The goal of video inpainting is to predict a completed video $\widehat{\textbf{\emph{Y}}}=\{\widehat{\textbf{\emph{y}}}_1,\widehat{\textbf{\emph{y}}}_2 ,\dots,\widehat{\textbf{\emph{y}}}_T\}$, which should be spatially and temporally consistent with the ground truth video $\textbf{\emph{Y}}=\{\textbf{\emph{y}}_1, \textbf{\emph{y}}_2,\dots, \textbf{\emph{y}}_T\}$.  Specifically, a mapping function from the input $\textbf{\emph{X}}$ to the output $\widehat{\textbf{\emph{Y}}}$ needs to be learned so that the conditional distribution $p(\widehat{\textbf{\emph{Y}}}|\textbf{\emph{X}})$ is approximate to $p(\textbf{\emph{Y}} | \textbf{\emph{X}})$.

Based on the fact that the contents of the corrupted regions in one frame may exist in neighboring frames, the video inpainting task can be formulated as a conditional pixel prediction problem:
\begin{equation}
p(\widehat{\textbf{\emph{Y}}}|\textbf{\emph{X}})=\prod_{t=1}^{T}
p(\widehat{\textbf{\emph{y}}}_t|\textbf{\emph{X}}_{r},\textbf{\emph{x}}_{t},\textbf{\emph{m}}_{t}),
\label{VIF}
\end{equation}
where $\textbf{\emph{X}}_{r}=\{\textbf{\emph{x}}_{t-n},\dots,\textbf{\emph{x}}_{t-1},\textbf{\emph{x}}_{t+1},\dots,\textbf{\emph{x}}_{t+n}\}$ 
%denotes a short clip of neighboring frames with a center moment $t$ and a temporal radius $n$
, namely \emph{reference frames}; $\textbf{\emph{x}}_{t}$ 
presents the current frame that needs to be inpainted, namely \emph{target frame}. 
Formally, the existing video inpainting methods aim to model the conditional distribution $p(\widehat{\textbf{\emph{y}}}_t|\textbf{\emph{X}}_{r},\textbf{\emph{x}}_{t},\textbf{\emph{m}}_{t})$ by training a deep neural network $D$, \emph{i.e.}, $\widehat{\textbf{\emph{y}}}_t=D(\textbf{\emph{X}}_{r},\textbf{\emph{x}}_{t},\textbf{\emph{m}}_{t})$.
The final output $\widehat{\textbf{\emph{Y}}}$ is obtained by processing the video frame by frame in temporal order.
Unfortunately, for our semi-supervised video inpainting setting, only the mask $\textbf{\emph{m}}_i$ of the one frame $\textbf{\emph{x}}_{i}$ is provided\footnote{Here, we usually assume that the annotated frame is the first frame of the video.}, while the masks of corrupted regions in other frames are unknown. This means that we don't know where other frames need to be inpainted except for the frames of the given mask.

%, i.e., it is unknown where those frames need to be completed. 
%Nevertheless, this fully-supervised setting usually assumes that all masks $\textbf{\emph{M}}$ are known, so we need to elaborately annotate the corrupted regions of each frame in the video, which is labor-intensive and expensive for real-world applications. 

%scc0812:从这里你才开始介绍你的semo
%我觉得哈，你应该把这部分提前，提前之后呢，再按照前面写的把公式（1）表述出来，这样的话，就是你的setting下的训练目标了
%\scc{To reduce the annotation cost}, we formulate a new semi-supervised video inpainting setting. In our semi-supervised setting, only the mask $\textbf{\emph{m}}_1$ of the first frame $\textbf{\emph{x}}_{1}$ is provided, while the masks $\textbf{\emph{M}}_{2}^{T}=\{\textbf{\emph{m}}_2, \dots,\textbf{\emph{m}}_T\}$ of corrupted regions in subsequent frames $\textbf{\emph{X}}_{2}^{T}=\{\textbf{\emph{x}}_2, \dots,\textbf{\emph{x}}_T\}$ are unknown, 
%scc0812:这句话重新写一下吧
%\scc{i.e. it is not known where to be inpainted in subsequent frames $\textbf{\emph{X}}_{2}^{T}$. }
%scc0812：下面这段可以在问题定义之后，再单列一段，说明一些对于你定义的这个问题的具体解决方法。

To tackle this issue, we decompose the semi-supervised video inpainting task into a pair of dual tasks: \emph{frame completion} and \emph{mask prediction}. The former aims to generate what to inpaint, while the latter is designed to estimate where to inpaint (via masks). 
%scc0812:
By this means, we can generate complete the current frame, and generate the 
corrupted regions mask of the next frame as well.
% Such processing can generate the corrupted regions mask of the next frame while completing the current frame. 
Thereafter,
by iterating frame by frame,
%scc0812：
we can complete the corrupted regions of the whole video sequence based on the only one known mask.
% we can only annotate the mask of one frame to complete the corrupted regions of the whole video sequence. 
Conditioned on a corrupted video sequence $\textbf{\emph{X}}$, 
the dual tasks 
% are
can be defined as:
%scc0812: 我把frame 和 mask 首字母大写了
\begin{itemize}
  \item \textbf{Frame completion} aims to
  learn 
  % the 
  a mapping $\bm\Phi$ to generate the completed frame $\widehat{\textbf{\emph{y}}}_{t}$ 
  % corresponded to 
of the target frame $\textbf{\emph{x}}_{t}$ utilizing the reference frames $\textbf{\emph{X}}_{r}$,
%\scc{Xr also explain}, 
\emph{i.e.}, 
  $\widehat{\textbf{\emph{y}}}_{t} = \bm \Phi(\textbf{\emph{X}}_{r},\textbf{\emph{x}}_{t},\textbf{\emph{m}}_{t})$;
  \item \textbf{Mask prediction} 
  expect to
  % learns
   learn a mapping $\bm\Psi$ to inversely generate mask $\textbf{\emph{m}}_{t+1}$ by using the frame $\textbf{\emph{x}}_{t+1}$ and the completed frame $\widehat{\textbf{\emph{y}}}_{t}$, \emph{i.e.}, 
  $\textbf{\emph{m}}_{t+1} = \bm \Psi( \textbf{\emph{x}}_{t+1},\widehat{\textbf{\emph{y}}}_{t})$.
\end{itemize}
\subsection{Network Design}
We design an end-to-end trainable framework to tackle semi-supervised video inpainting task. As shown in Fig.\ref{Fig2}, our framework consists of a \textbf{completion network} and a \textbf{mask prediction network}. The former aims to use the temporal information of the reference frames $\textbf{\emph{X}}_{r}$ to complete the corrupted regions $\textbf{\emph{m}}_{t}$ of the target frame $\textbf{\emph{x}}_{t}$, 
while the latter 
% uses
utilizes
the completed target frame $\widehat{\textbf{\emph{y}}}_{t}$ to predict the corrupted regions $\textbf{\emph{m}}_{t+1}$ of the subsequent frame $\textbf{\emph{x}}_{t+1}$.
%scc0812: 你再检查一下这两篇中单个词占用一行的情况，想办法避免一下
\subsubsection{Completion Network.}
As shown in Fig.\ref{Fig2}, completion network consists of four parts: frame-level encoder, feature alignment module, feature aggregation module,
and frame-level decoder. The frame-level encoder aims to extract deep features from low-level pixels of each frame, while the frame-level decoder is used to decode completed deep features into the frame. 
They consist of multiple convolutional layers and residual blocks with ReLUs as the activation functions. 
\textbf{\emph{Feature alignment}} and \textbf{\emph{feature aggregation}} 
modules 
are the core components of the completion network. The former performs reference frame alignment at the feature level to eliminate image changes between the reference frame and the target frame, 
while 
the latter aggregates the aligned reference frame features to complete corrupted regions of the target frame.

\noindent\textbf{\emph{Feature Alignment.}} 
Due to the 
image variation caused by camera and object motion, it is difficult to directly utilize the temporal information of the reference frames $\textbf{\emph{X}}_{r}$ to complete the corrupted regions $\textbf{\emph{m}}_{t}$ of the target frame $\textbf{\emph{x}}_{t}$. Therefore, an extra alignment module is necessary for video inpainting. 
Notably, 
%scc:0812 你看看下面注释掉的句子有没有必要存在
% to improve the complex geometric transformation ability, 
deformable convolution~\cite{8237351} can obtain information away from its regular local neighborhood by learning offsets of the sampling convolution kernels.
% , so as to improve the complex geometric transformation ability. 
Motivated by the capacity of deformable convolution, a Deformable Convolution Alignment (DCA) block is 
% used 
designed to perform reference 
% frames 
frame alignment at the feature level. Specifically, 
%scc0812:这句话没有说完整
for the target frame feature $\textbf{\emph{f}}_{t}$ and the reference frame feature $\textbf{\emph{f}}_{r}$ extracted by the frame-level encoder, we first cascade them 
%where $r\in\{t-n,\dots, t-1, t+1, \dots, t+n\}$.
%First of all, they are cascaded 
to predict the offsets $\theta=\{\triangle{\textbf{\emph{p}}_{n}}|n=1,\ldots,|R|\}$ of the convolution kernel, where $r\in\{t-n,\dots, t-1, t+1, \dots, t+n\}$, and $R=\{(-1,-1),(-1,0),\ldots,(1,1)\}$ denotes a regular grid of a $3\times3$ kernel. Then, with the predicted offset $\theta$, the aligned feature $\textbf{\emph{f}}_{r}^{a}$ of the reference frame feature $\textbf{\emph{f}}_{r}$ is obtained by a deformable convolution layer $\mathcal{DCN}$:
\begin{equation}
\textbf{\emph{f}}_{r}^a = \mathcal{DCN}(\textbf{\emph{f}}_{r}, \theta).
\label{fdc}
\end{equation}

In practice, to facilitate feature alignment of the reference frames more accurately, we cascade four DCA blocks in the feature alignment module to enhance its transformation ﬂexibility and capability. Ablation study on the performance of the feature alignment with different numbers of the DCA blocks can be found in Section.~\ref{Ablation Studies}.
%Section.\ref{Ablation Studies}.

\noindent\textbf{\emph{Feature Aggregation.}}
\begin{figure}[tb]
\centering%height=3.8cm,width=17.5cm
\includegraphics[scale=0.32]{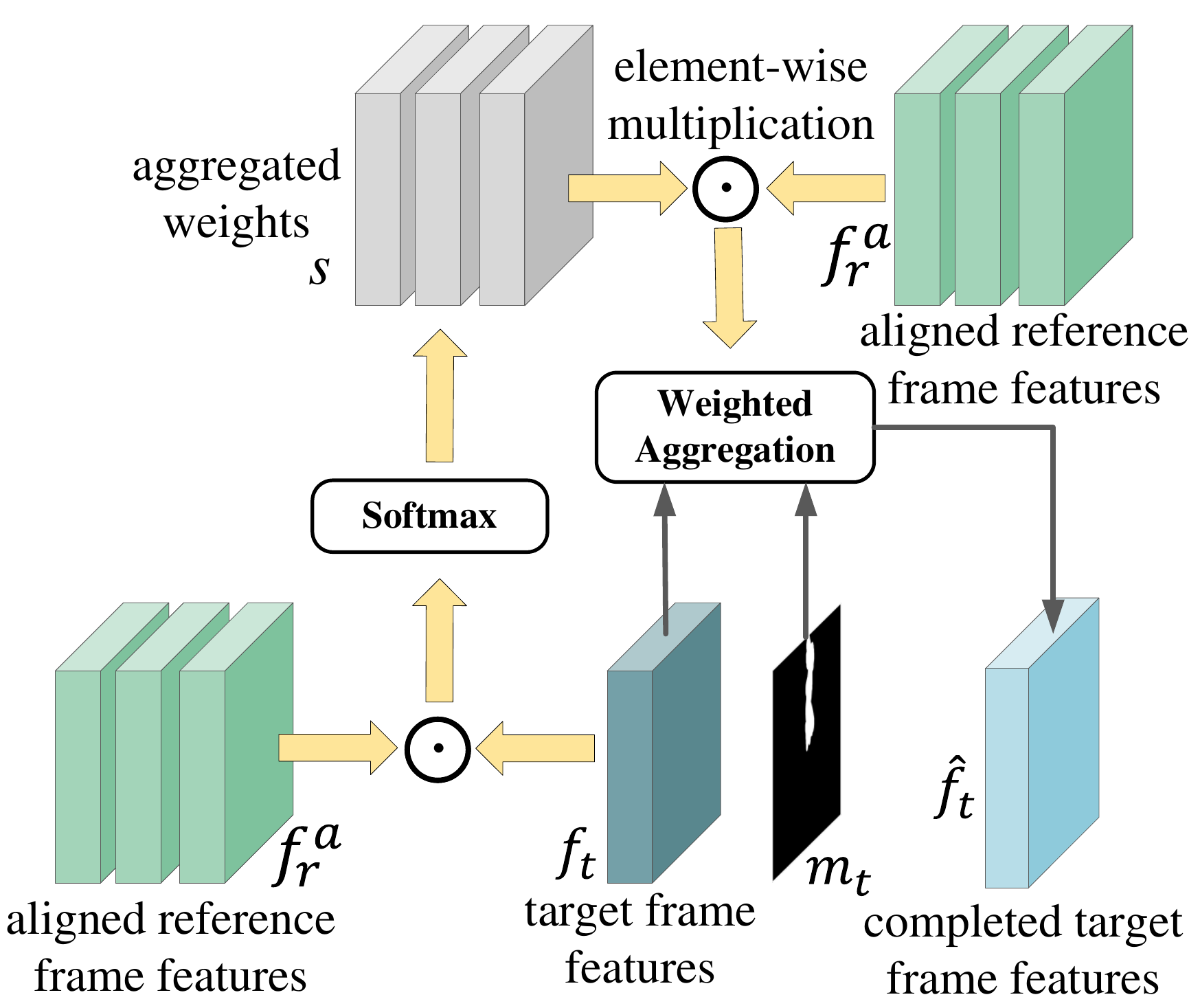}
\vspace{-0.2cm}
\caption{Illustration of adaptive temporal features aggregation module.}
\label{FigC}
\vspace{-0.5cm}
\end{figure}
Due to occlusion, blur and parallax problems, different aligned reference frame features are not equally beneficial for 
the
reconstruction of corrupted contents of the target frame. Therefore, an adaptive temporal features aggregation module is 
% used 
introduced to dynamically aggregate aligned reference frame features as shown in Fig.\ref{FigC}. 
Specifically, for each aligned reference frame feature $\textbf{\emph{f}}_{r}^a$, we first calculate the aggregate weights $\textbf{\emph{s}}_{r}$ by the softmax function:
\begin{equation}
\textbf{\emph{s}}_{r}=\frac{exp\left({\mathcal{Q}(\textbf{\emph{f}}_{t})}^T\cdot\mathcal{K}(\textbf{\emph{f}}_{r}^{a})\right)}{\sum_r{exp\left({\mathcal{Q}(\textbf{\emph{f}}_{t})}^T\cdot\mathcal{K}(\textbf{\emph{f}}_{r}^{a})\right)}},
\label{softmax}
\end{equation}
where $\mathcal{Q}(\cdot)$ and $\mathcal{K}(\cdot)$ denote $1\times1$ 2D convolution. 
After obtaining the aggregated weight $\textbf{\emph{s}}$ for all reference frame features, the modulation feature $\textbf{\emph{h}}_{r}$ corresponding to the feature $\textbf{\emph{f}}_{r}^a$ is obtained by a pixel-wise manner: 
%scc0812:不要写在句子中，写成单个公式的形式。
\begin{equation}
\textbf{\emph{h}}_{r}=\mathcal{V}(\textbf{\emph{f}}_{r}^{a})\odot\textbf{\emph{s}}_{r},
\label{softmax1}
\end{equation}
%\scc{$\textbf{\emph{h}}_{r}=\mathcal{V}(\textbf{\emph{f}}_{r}^{a})\odot\textbf{\emph{s}}_{r}$}, 
where $\mathcal{V}(\cdot)$ denotes $1\times1$ 2D convolution, $\odot$ denotes the element-wise multiplication. Finally, the aggregated features $\widehat{\textbf{\emph{f}}}_{t}$ are obtained by a fusion convolutional layer:
\begin{equation}
\widehat{\textbf{\emph{f}}}_{t}=\mathcal{{A}}([\textbf{\emph{h}}_{t-n},\ldots,\textbf{\emph{h}}_{t+n},\textbf{\emph{f}}_{t},\textbf{\emph{m}}_{t}]),
\label{fusion}
\end{equation}
where $[\cdot,\cdot,\cdot]$ denotes the concatenation operation, $\mathcal{{A}}$ is a $1\times1$ convolution layer. The final inpainted target frame $\widehat{\textbf{\emph{y}}}_{t}$ corresponding to the target frame $\textbf{\emph{x}}_{t}$ can be obtained by decoding $\widehat{\textbf{\emph{f}}}_{t}$ with the frame-level decoder. 

\subsubsection{Mask Prediction Network.}
Mask prediction network aims to predict the corrupted regions of the video frame. A naive idea is to use the corrupted regions as a segmentation object to generate corrupted regions mask for subsequent frames by state-of-the-art video object segmentation (VOS) methods. Although VOS methods have achieved significant results in the past few years, their performance is unsatisfactory for segmenting corrupted regions of videos
due to the following reasons:
% The reasons for this result are: 
1) the appearance of the corrupted regions varies greatly in the video; 2) the boundary between the corrupted regions and the uncorrupted regions is blurred. 
%scc0812:我删除了下面这句话
% These are exactly the problems that current VOS methods need to solve urgently. 
Fortunately, we 
% found
%\scc{find that the difference between the video frames before and after completion is only the corrupted regions of the current frame.}
find the fact that the only difference between the video frames before and after completion is the corrupted regions of the current frame.%scc0812: maybe rewrite like: we find the fact that the only difference .....
Therefore, we design a mask prediction network that utilizes current completed frame $\widehat{\textbf{\emph{y}}}_{t}$ and subsequent frame $\textbf{\emph{x}}_{t+1}$ as 
% inputs
input to generate corrupted region mask $\textbf{\emph{m}}_{t+1}$ for subsequent frame.

As shown in Fig.\ref{Fig2}, mask prediction network consists of three parts: frame-level encoder, feature alignment module 
, and frame-level decoder. 
Specifically, we first extract the deep features $\textbf{\emph{g}}_{t}$ and $\textbf{\emph{f}}_{t+1}$ of the current completed frame $\widehat{\textbf{\emph{y}}}_{t}$ and subsequent frame $\textbf{\emph{x}}_{t+1}$ by the frame-level encoder, respectively. Here, to enrich the deep feature $\textbf{\emph{g}}_{t}$ of the completed frame $\widehat{\textbf{\emph{y}}}_{t}$, we concatenate the completed feature $\widehat{\textbf{\emph{f}}}_{t}$ obtained in the completion network with $\textbf{\emph{g}}_{t}$, i.e.,  $\widehat{\textbf{\emph{q}}}_{t}=[\textbf{\emph{g}}_{t},\widehat{\textbf{\emph{f}}}_{t}]$. Then, the feature alignment module is used to align the concatenated feature $\widehat{\textbf{\emph{q}}}_{t}$, which aims to eliminate the effects of image changes between current completed frame $\widehat{\textbf{\emph{y}}}_{t}$ and subsequent frame $\textbf{\emph{x}}_{t+1}$. Note that the feature alignment module of mask prediction network has the same structure as the feature alignment module of completion network, and the parameters between them are shared. Finally, the aligned feature $\widehat{\textbf{\emph{q}}}_{t}^{a}$ and subsequent frame feature $\textbf{\emph{f}}_{t+1}$ are concatenated into a frame-level decoder to generate the corrupted regions mask $\textbf{\emph{m}}_{t+1}$ for subsequent frame $\textbf{\emph{x}}_{t+1}$. Details of the frame-level encoder and frame-level decoder can be found in the supplementary materials.

\subsection{Cycle Consistency}
\begin{figure}[tb]
\centering%height=3.8cm,width=17.5cm
\includegraphics[scale=0.43]{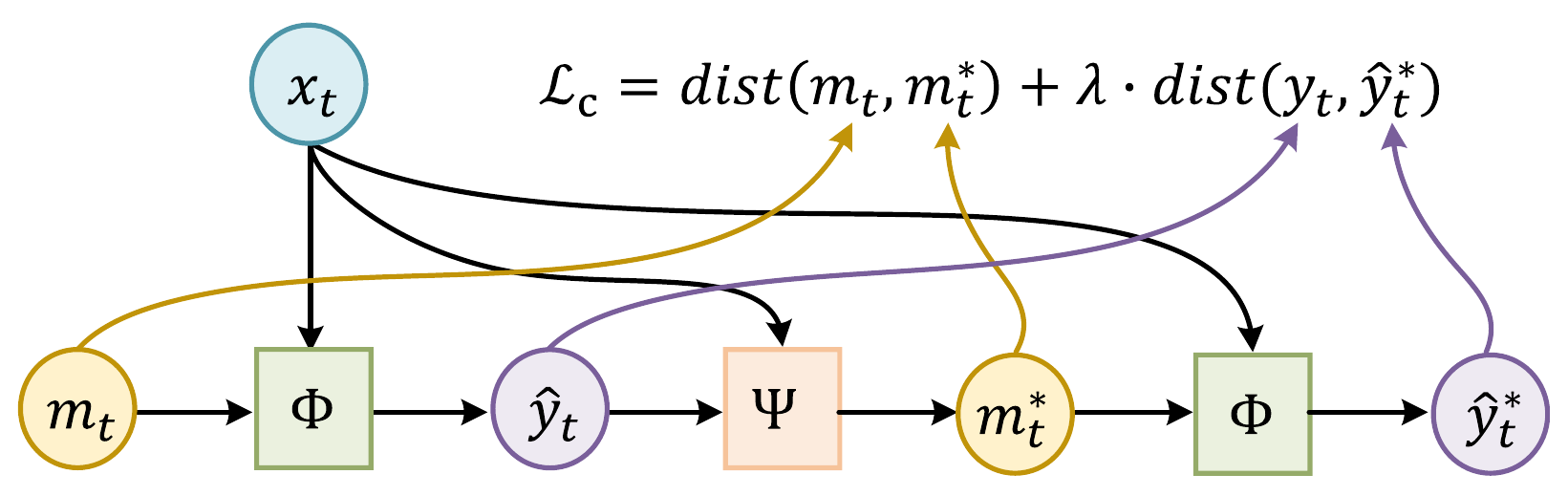}
\vspace{-0.2cm}
\caption{Illustration of cycle consistency.}
\label{FigC1}
\vspace{-0.3cm}
\end{figure}
In our task, the correspondence between $\widehat{\textbf{\emph{y}}}_{t}$ and $\textbf{\emph{m}}_{t}$ is one-to-one, so the frame completion and mask prediction exist simultaneously.
In other words,
% \emph{i.e.},  %scc0812
for video sequence $\textbf{\emph{X}}$, given the corresponding mask $\textbf{\emph{m}}_{t}$ of target frame $\textbf{\emph{x}}_{t}$, the completed frame $\widehat{\textbf{\emph{y}}}_{t}$ can be obtained by $\widehat{\textbf{\emph{y}}}_{t}=\Phi(\textbf{\emph{X}}_{r},\textbf{\emph{x}}_{t},\textbf{\emph{m}}_{t})$. Conversely, given the corresponding completed frame $\widehat{\textbf{\emph{y}}}_{t}$ of target frame $\textbf{\emph{x}}_{t}$, the mask $\textbf{\emph{m}}_{t}$ can be obtained by $\textbf{\emph{m}}_{t}=\Psi(\textbf{\emph{x}}_{t},\widehat{\textbf{\emph{y}}}_{t})$. 
Ideally, if the mapping $\Phi$ and $\Psi$ 
both can 
capture accurate correspondence, nesting them together can obtain the following cycle consistency: 
\begin{equation}
\widehat{\textbf{\emph{y}}}_{t}^{\ast}=\Phi(\textbf{\emph{X}}_{r},\textbf{\emph{x}}_{t},\Psi(\textbf{\emph{x}}_{t},\widehat{\textbf{\emph{y}}}_{t})), 
\label{Y_cycle}
\end{equation}
\begin{equation}
\textbf{\emph{m}}_{t}^{\ast}=\Psi(\textbf{\emph{x}}_{t},\Phi(\textbf{\emph{X}}_{r},\textbf{\emph{x}}_{t},\textbf{\emph{m}}_{t})).
\label{M_cycle}
\end{equation}
%where $n\geq1$ denote the number of iterations. $\widehat{\textbf{\emph{Y}}}_{t}^{(n)}$ and $\textbf{\emph{M}}_{t}^{(n)}$ denote the results after $nth$ iterations. if $n=1$, then $\widehat{\textbf{\emph{Y}}}_{t}^{(0)}=\widehat{\textbf{\emph{Y}}}_{t},\textbf{\emph{M}}_{t}^{(0)}=\textbf{\emph{M}}_{t}$.

Eq.(\ref{Y_cycle}) and Eq.(\ref{M_cycle}) give us the solution to generate reliable and consistent correspondence between the mapping $\Phi$ and $\Psi$ by formulating the loss as
follows,
\begin{equation}
\mathcal{{L}}_{y}=dist(\widehat{\textbf{\emph{y}}}_{t},\widehat{\textbf{\emph{y}}}_{t}^{\ast}),
\label{L-cycle_Y}
\end{equation}
\begin{equation}
\mathcal{{L}}_{m}=dist(\textbf{\emph{m}}_{t}, \textbf{\emph{m}}_{t}^{\ast}),
\label{L-cycle_M}
\end{equation}
where $dist(\cdot,\cdot)$ is $L_1$ distance function. 

As shown in Fig.\ref{FigC1}, by combining Eq.(\ref{L-cycle_Y}) and Eq.(\ref{L-cycle_M}), we obtain the hybrid loss as
follows,
\begin{equation}
\mathcal{{L}}_{c}=\mathcal{{L}}_{m}+\lambda_{y}\mathcal{{L}}_{y},
\label{L-cycle}
\end{equation}
where $\lambda_{y}$ is the 
non-negative 
trade-off parameter. 

\subsection{Loss Function}
We train our network by minimizing the following loss:
\begin{equation}
\begin{aligned}
\mathcal{{L}}=\lambda_{f}\mathcal{{L}}_{f}+\lambda_{s}\mathcal{{L}}_{s}+\lambda_{c}\mathcal{{L}}_{c},
\label{L-total}
\end{aligned}
\end{equation}
where $\mathcal{{L}}_{f}$ is the frame reconstruction loss, $\mathcal{{L}}_{s}$ is the mask prediction loss and $\mathcal{{L}}_{c}$ is the cycle consistency loss. $\lambda_{f}$, $\lambda_{s}$ and $\lambda_{c}$ are the trade-off parameters. In 
real
implementation, we empirically set the weights of different losses as: $\lambda_{f}=2.5$, $\lambda_{s}=0.25$ and $\lambda_{c}=1$. More details of 
the analysis of %scc0812
loss function 
% are shown 
can be found in supplementary materials.

\begin{figure}[tb]
\centering%height=3.0cm,width=15.5cm
\includegraphics[height=3cm,width=8.2cm]{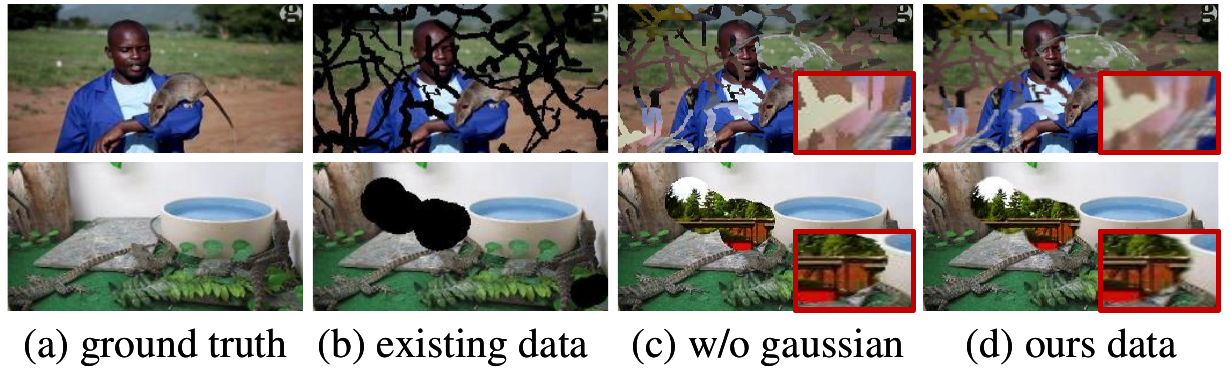}
\vspace{-0.3cm}
\caption{Example of generated traning data.}
\label{Fig_D}
\vspace{-0.5cm}
\end{figure}

\section{Experiments}
\label{Experiments}
\subsection{Dataset and Implementation Details}
\noindent\textbf{Training Data Generation.}
% The diversity of the training set is the essential prerequisite for whether a network model can be robust to various possible video corrupts.
%scc0812:
In general, 
the diversity of the training set is the essential prerequisite for the robustness of the trained model.
Existing
benchmark 
datasets are usually generated by ${\textbf{\emph{x}}_i}=(1-{\textbf{\emph{m}}_i})\odot{\textbf{\emph{y}}_i}+{\textbf{\emph{m}}_i}\odot{\textbf{\emph{u}}_i}$, where ${\textbf{\emph{u}}_i}$ is a noise signal 
% of 
having the same size as ${\textbf{\emph{y}}_i}$, $\odot$ is the element-wise multiplication. Traditional fully-supervised video inpainting tasks usually assume ${\textbf{\emph{u}}_i}$ as a constant value or certain kind of noise (Fig.\ref{Fig_D}(b)). 
% However, such a setting is not feasible for our semi-supervised video inpainting task, since setting ${\textbf{\emph{u}}_i}$ as a constant value or a certain kind of noise would make it and ${\textbf{\emph{m}}_i}$ easy to be distinguished by a deep neural net or even a simple linear classifier from a natural image. 
% Such a situation converts the semi-supervised video inpainting task into a fully-supervised video inpainting task with almost perfect prediction of ${\textbf{\emph{m}}_i}$, which can be solved using existing methods, but such an assumption generally does not hold in real-world scenarios.
%scc0812: 你看一下我改的这段是不是你想表达的意思
However, when ${\textbf{\emph{u}}_i}$ is set as a constant value or a certain kind of noise, both ${\textbf{\emph{u}}_i}$ and ${\textbf{\emph{m}}_i}$ would be easily distinguished by a deep neural net or even a simple linear classifier from a natural image.
In this way, video inpainting task is fallen into the 
fully-supervised category with almost perfect prediction of ${\textbf{\emph{m}}_i}$,
which can be solved using existing methods.
% but such an assumption generally does not hold in real-world scenarios.
In fact, such an assumption generally does not hold in real-world scenarios.

Therefore,
for semi-supervised video inpainting tasks, the key to dataset generation is how to define noise ${\textbf{\emph{u}}_i}$ to mask it as different as possible from ${\textbf{\emph{x}}_i}$ in image pattern. In this paper, we use real-world image patches to define noisy ${\textbf{\emph{u}}_i}$. This definition can ensure that the local patches between the noise  ${\textbf{\emph{u}}_i}$ and the image ${\textbf{\emph{x}}_i}$ cannot be easily distinguished, enforcing the network to infer the location ${\textbf{\emph{m}}_i}$ of the noise ${\textbf{\emph{u}}_i}$ according to the context of the ${\textbf{\emph{x}}_i}$, which eventually improves the generalization ability for real-world data. Furthermore, it is worth noting that the existing methods of fully supervised video inpainting usually use the fixed-shape ${\textbf{\emph{m}}_i}$ (\emph{e.g,} rectangle) to generate datasets, which is disadvantageous for our semi-supervised video inpainting task. This is because the fixed-shape ${\textbf{\emph{m}}_i}$ introduces prior knowledge into the training of the model, which encourages the mask prediction network to locate corrupted regions based on the mask shape. Therefore, free-form strokes~\cite{chang2019free} with shape diversity are used as our ${\textbf{\emph{m}}_i}$ to generate the dataset, making the mask prediction network harder to infer corrupted regions by using shape information. 

\begin{table*}[!t]
\caption{Quantitative results of video inpainting. The term $\emph{Semi}$ denotes `Semi-supervised' for short, $\emph{Our\_ComNet}$ represents the completion network trained in the fully supervised manner.}
\renewcommand\arraystretch{0.5}
  \centering
  \scriptsize
    \begin{tabular}{c|c|cccc|cccc}
    \hline
          &\multicolumn{1}{c|}{} & \multicolumn{4}{c|}{Youtube-vos}      & \multicolumn{4}{c}{DAVIS} \\
    \hline
          & Semi & PSNR$\uparrow$ & SSIM$\uparrow$ &  $E_{warp}\downarrow$  &LPIPS$\downarrow$  & PSNR$\uparrow$ & SSIM$\uparrow$ &  $E_{warp}\downarrow$  &LPIPS$\downarrow$  \\
    \hline
    TCCDS &\XSolidBrush &23.418&0.8119 &0.3388&1.9372 &28.146&0.8826   &0.2409&1.0079  \\
    VINet &\XSolidBrush &26.174 &0.8502 &0.1694 &1.0706 &29.149 &0.8965 &0.1846 &0.7262 \\
    DFVI  &\XSolidBrush &28.672 &0.8706 &0.1479 &0.6285 &30.448&0.8961    &0.1640&0.6857 \\
    CPVINet &\XSolidBrush  &28.534    &0.8798&0.1613&0.8126&30.234&0.8997 &0.1892&0.6560  \\
    FGVC  &\XSolidBrush &24.244&0.8114 &0.2484&1.5884&28.936&0.8852   &0.2122&0.9598 \\
    OPN   &\XSolidBrush &30.959&0.9142 &0.1447&0.4145 &32.281&0.9302     &0.1661&0.3876\\
    STTN  &\XSolidBrush &28.993&0.8761 &0.1523&0.6965&28.891&0.8719   &0.1844&0.8683 \\
    FuseFormer&\XSolidBrush &29.765&0.8876 &0.1463&0.5481&29.627&0.8852&0.1767&0.6706 \\
    E2FGVI &\XSolidBrush &30.064&0.9004&0.1490&0.5321  &31.941&0.9188&0.4579&0.6344 \\
    Our\_ComNet &\XSolidBrush &\textbf{31.291} &\textbf{0.9237} &\textbf{0.1423} &\textbf{0.3918} &\textbf{32.807}&\textbf{0.9401}&\textbf{0.1503}&\textbf{0.3681}\\
    \hline
    RANet+OPN  &\Checkmark &21.826 &0.8058 &0.2446 &0.9346 &20.609 &0.8094 &0.3251 &0.7315\\
    STM+FuseFormer &\Checkmark &23.371 &0.8283 &0.1919 &0.8613 &19.917 &0.8906 &0.2569 &0.8119\\
    GMN+STTN &\Checkmark &22.680 &0.8145 &0.1822 &0.7335 &20.394 &0.8204 &0.2647 &0.8071\\
    HMMN+E2FGVI &\Checkmark &25.255 &0.8539 &0.1803 &0.7019 &22.696 &0.8635 &0.2429 &0.7313\\
    Ours   &\Checkmark &\textbf{30.834}&\textbf{0.9135}&\textbf{0.1452}&\textbf{0.4171}&\textbf{32.097}&\textbf{0.9263}&\textbf{0.1534}&\textbf{0.3852}\\
    \hline
    \end{tabular}%
  \label{tab_Q}%
  \vspace{-0.2cm}
\end{table*}%

In addition, the mixed image ${\textbf{\emph{x}}_i}$ obtained directly through ${\textbf{\emph{x}}_i}=(1-{\textbf{\emph{m}}_i})\odot{\textbf{\emph{y}}_i}+{\textbf{\emph{m}}_i}\odot{\textbf{\emph{u}}_i}$ using ${\textbf{\emph{y}}_i}$ and ${\textbf{\emph{u}}_i}$ would lead to noticeable edges (Fig.\ref{Fig_D}(c)), which are strong indicators for distinguishing corrupted regions. Such edge priors will inevitably sacrifice the semantic understanding capability of the mask prediction network. Therefore, we use iterative Gaussian smoothing~\cite{wang2018image} to extend ${\textbf{\emph{m}}_i}$ in the process of dataset generation, and employ alpha blending in the contact regions between ${\textbf{\emph{u}}_i}$ and ${\textbf{\emph{y}}_i}$, which effectively avoids introducing obvious edge priors into the dataset. Our dataset is generated on the basis of Youtube-vos~\cite{Xu2018YouTube}, \emph{i.e.} video frames in Youtube-vos as our ${\textbf{\emph{y}}_i}$. Following the settings of the original Youtube-vos dataset, our generated dataset contains 3471, 474, and 508 video clips in training, validation, and test sets, respectively.

\noindent\textbf{Testing Dataset}
To evaluate the effectiveness of our method, two popular video object segmentation datasets are taken to evaluate our model, including Youtube-vos~\cite{Xu2018YouTube} and DAVIS~\cite{Perazzi2016A}. Following the previous settings~\cite{Li2020ShortTermAL,xu2019deep}, 
% the test sets 
% the 
DAVIS 
% dataset
contains 60 video clips. To ensure the comparability of experimental results, all baseline methods used for comparison are fine-tuned multiple times on our generated dataset 
by their released models and codes, and the best results are reported.% in this paper.

\begin{figure}[tb]
\centering%height=3.0cm,width=15.5cm
\includegraphics[height=7.6cm,width=8.2cm]{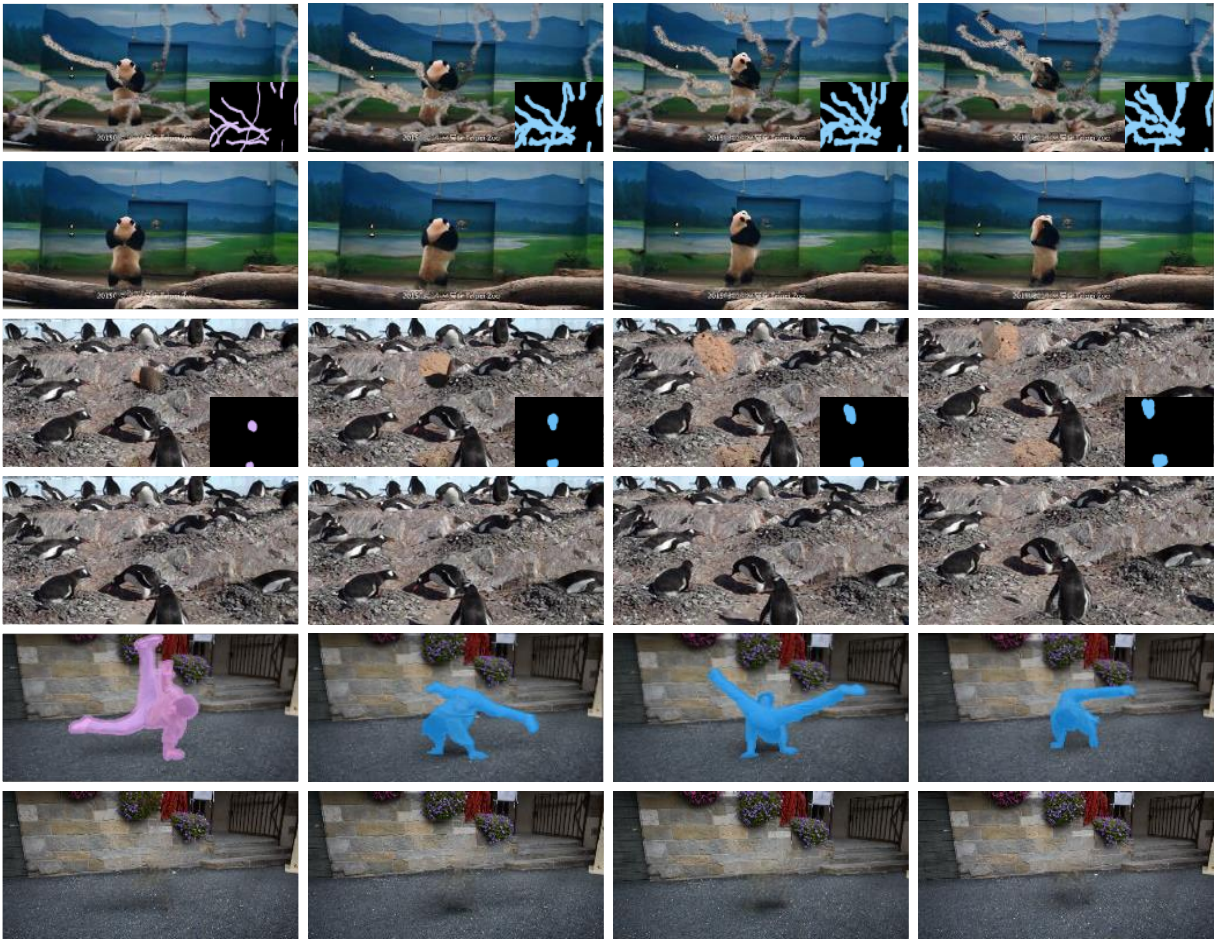}
\vspace{-0.25cm}
\caption{Three example of inpainting results with our method. The top row shows sample frames with the mask, where pink denotes the manually annotated object mask, and blue denotes the segmentation mask generated by the model. The completed results are shown in the bottom row.}
\label{Fig6}
\vspace{-0.45cm}
\end{figure}

\noindent\textbf{Implementation Details.} We use PyTorch to implement our model. The proposed semi-supervised video inpainting network is trained by using Adam optimizer with learning rate 1e-4 and $\beta=(0.9,0.999)$. 
% The video sequences are resized during the training to $256\times256$ as inputs. 
The video sequences are resized to $256\times256$ during the training. 
More training details are provided in our supplementary materials.

\subsection{Video Inpainting Evaluation}
\noindent\textbf{Baselines and evaluation metrics.}
To evaluate the video inpainting ability of our model, nine state-of-the-art video inpainting methods are used as our baselines, including one optimization-based method: TCCDS~\cite{huang2016temporally} and eight learning-based methods: VINet~\cite{Kim_2019_CVPR,8931251}, DFVI~\cite{xu2019deep}, CPVINet~\cite{lee2019cpnet}, FGVC~\cite{Gao-ECCV-FGVC}, OPN~\cite{9010390}, STTN~\cite{yan2020sttn}, FuseFormer~\cite{liu2021fuseformer}, and E2FGVI~\cite{li2022towards}. Note that there was no work focusing on semi-supervised video inpainting task before, 
% so 
% these 
baselines mentioned above
are all 
% were 
worked in a fully-supervised manner. Furthermore, to strengthen the baselines in the semi-supervised setting, we also introduce the methods of connecting VOS and VI as our baselines, \emph{i.e.}, using the existing VOS methods (RANet~\cite{Wang_2019_ICCV}, GMN~\cite{lu2020video}, STM~\cite{9008790}, and HMMN~\cite{seong2021hierarchical}) to obtain full mask information of the video, then completing missing regions of the video with the fully-supervised VI methods (OPN~\cite{9010390}, STTN~\cite{yan2020sttn}, FuseFormer~\cite{liu2021fuseformer}, and E2FGVI~\cite{li2022towards}). The quantitative results of video inpainting are reported by four metrics, \emph{i.e.},  PSNR~\cite{9008384}, SSIM~\cite{9010390}, LPIPS~\cite{zhang2018unreasonable}, and flow warping error $E_{warp}$~\cite{lai2018learning}.
\begin{figure}[tb]
\centering%height=3.0cm,width=15.5cm
\includegraphics[height=4cm,width=8.2cm]{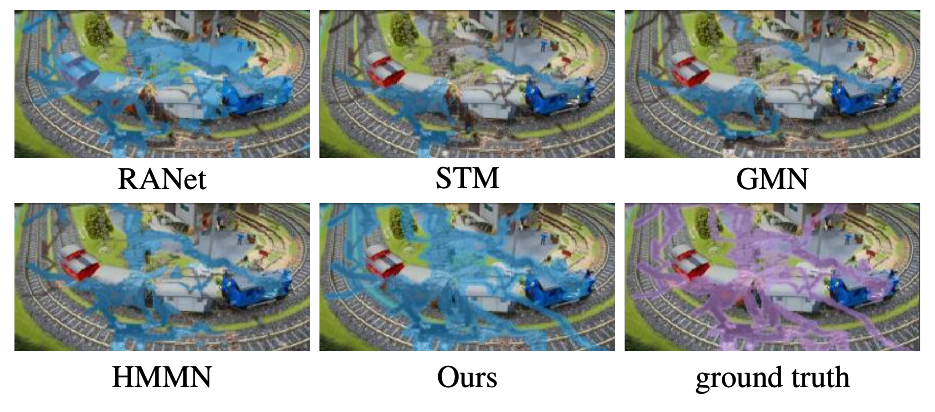}
\vspace{-0.3cm}
\caption{Example of corrupted regions segmentation.}
\label{Fig_s}
\vspace{-0.3cm}
\end{figure}

\noindent\textbf{Experimental results and analysis.}
The quantitative results of video inpainting on Youtube-vos and DAVIS datasets are summarized in Tab.\ref{tab_Q}. It can be seen from Tab.\ref{tab_Q} that the PSNR, SSIM, $E_{warp}$ and LPIPS of our semi-supervised model are comparable to the fully-supervised baselines on two datasets, and significantly outperform the method concatenating VOS and VI. Fig.\ref{Fig6} shows some visual results of semi-supervised video inpainting obtained by our method. It can be seen from Fig.\ref{Fig6} that the proposed method can generate the spatio-temporally consistent content in the missing regions of the video. It should be noted that our semi-supervised model only uses the mask of the first frame to indicate corrupted regions when completing the whole video, 
which is 
greatly
beneficial for the practical application of video inpainting compared to the fully-supervised baselines that requires annotations of all frames. More video inpainting results can be found in our supplementary materials.

\begin{table}[tb]
\caption{Evaluation results of mask prediction.}
  \centering
  \setlength\tabcolsep{5pt}
  \scriptsize
    \begin{tabular}{c|c c|c c}
    \hline
        & \multicolumn{2}{c|}{Youtube-vos} & \multicolumn{2}{c}{DAVIS} \\
    \hline
     & BCE$\downarrow$ & IOU$\uparrow$ & BCE$\downarrow$ & IOU$\uparrow$ \\
    \hline
    RANet & 4.053  & 0.411  & 4.112  & 0.381  \\
    GMN & 2.381  & 0.587  & 4.602  & 0.363  \\
    STM & 2.032  & 0.605  & 4.851  & 0.316  \\
    HMMN & 1.567  & 0.719  & 3.738  & 0.456  \\
    Ours & \bf{1.148}  & \bf{0.934}  &\bf{1.327}  &\bf{0.832}  \\
    \hline
    \end{tabular}%
    \vspace{-0.2cm}
  \label{tab mask}%
\end{table}%
\subsection{Mask Prediction Evaluation}
\noindent\textbf{Baselines and evaluation metrics.}
In this section, we use the learned mask prediction network to evaluate the corrupted regions segmentation ability of our model. 
To 
intuitively reflect the segmentation ability of our model, the specialized semi-supervised video object segmentation methods are used as baselines, including RANet~\cite{Wang_2019_ICCV}, GMN~\cite{lu2020video}, STM~\cite{9008790}, and HMMN~\cite{seong2021hierarchical}. 
In our experiment, these segmentation methods are fine-tuned multiple times on our generated dataset by their released models and codes, and the best results are reported in this paper. 
Furthermore, the quantitative results of mask segmentation are reported by two metrics, \emph{i.e.}, intersection over union (IOU) and binary cross entropy (BCE) loss.

\noindent\textbf{Experimental results and analysis.}
Different from the existing semi-supervised video object segmentation methods, the proposed mask prediction network use the completed current frame and the subsequent frame as input to generate the segmentation masks. Ideally, the difference between these two frames is only the corrupted regions. Tab.\ref{tab mask} lists the values of BCE and IOU of our method and the baselines on Youtube-vos~\cite{Xu2018YouTube} and DAVIS~\cite{7780454} datasets. It can be observed that our mask prediction network obtains the best results. Fig.\ref{Fig_s} shows segmentation results obtained by different methods. It can be seen from Fig.\ref{Fig_s} that the proposed mask prediction network can obtain the segmentation result closer to ground truth, while the segmentation results of RANet, GMN, STM, and HMMN lose part of the corrupted regions. 
%scc0812:不要说这是reason，改成这验证了 verify
%\scc{This is the main reason why the inpainting effect of the methods concatenating VOS and VI is unsatisfactory.}
This also verifies that the inpainting effect of the methods concatenating VOS and VI is unsatisfactory.

\subsection{Ablation Studies}
\label{Ablation Studies}
\noindent\textbf{Effectiveness of Complete Network.}
To verify the effectiveness of the proposed frame completion network, we test the completion network trained in a fully-supervised manner, and the testing results are shown in Tab.\ref{tab_Q}. Compared with the fully-supervised baselines, our model obtains the best performance. In addition, we also conduct semi-supervised video inpainting tests using the best three baseline models instead of our complete network. As shown in Tab.\ref{tab_c}, our semi-supervised model outperforms the other three combined models, demonstrating our completion network's superiority in video inpainting.

\begin{table}[tb]
  \centering
\scriptsize
  \renewcommand\tabcolsep{3pt}
  \caption{Ablation study of complete network and cycle consistency loss. $\emph{MPN}$ denotes mask prediction network.}
    \begin{tabular}{c|cccc}
    \hline
          & PSNR$\uparrow$ & SSIM$\uparrow$ &  $E_{warp}\downarrow$  &LPIPS$\downarrow$\\
    \hline
    OPN+MPN &29.149&0.9013&0.1506&0.4282\\
    OPN+MPN+$\mathcal{{L}}_{c}$ &30.017&0.9109&0.1490&0.4227\\
    \hline
    CPVINet+MPN &26.924&0.8696&0.1684&0.8161\\
    CPVINet+MPN+$\mathcal{{L}}_{c}$ &27.952&0.8765&0.1636&0.8138\\
    \hline
    E2FGVI+MPN &28.139&0.8857&0.1513&0.5521\\
    E2FGVI+MPN+$\mathcal{{L}}_{c}$ &29.721&0.8987&0.1501&0.5496\\
    \hline
    w/o $\mathcal{{L}}_{c}$&29.522&0.9001&0.1476&0.4235  \\
    final model  &{30.834}&{0.9135}&{0.1452}&{0.4171}  \\
    \hline
    \end{tabular}%
  \label{tab_c}%
  \vspace{-0.1cm}
\end{table}%

\begin{table}[tb]
\renewcommand\tabcolsep{4pt}
\scriptsize
\centering
\caption{Ablation study of mask annotation location.}
\begin{tabular}{ccccccc}
\hline 
{annotation location}   & PSNR$\uparrow$ & SSIM$\uparrow$ &  $E_{warp}\downarrow$  &LPIPS$\downarrow$  \\
\hline
{first frame}      &{30.834}&{0.9135}&{0.1452}&{0.4171}\\
{middle frame}      &{30.826}    &{0.9179}       &{0.1461}  &{0.4196}\\
{last frame}    &{30.796}&{0.9165}&{0.1443}&{0.4147}\\
\hline
\end{tabular}
\vspace{-0.3cm}
\label{al}
\end{table}

\noindent\textbf{Effectiveness of Cycle Consistency.}
The cycle consistency loss $\mathcal{{L}}_{c}$ is used to regularize the training parameters of both the completion network and mask prediction network. In this section, we verify the effectiveness of $\mathcal{{L}}_{c}$. As shown in Tab.\ref{tab_c}, the performance of the four models trained with $\mathcal{{L}}_{c}$ is improved compared to the model trained without $\mathcal{{L}}_{c}$. Therefore, 
% we think 
we can draw the conclusion that $\mathcal{{L}}_{c}$ can facilitate the complete network and mask prediction network to generate reliable and consistent correspondences, thereby improving the quality of inpainted videos.

%scc0814：ok
\noindent\textbf{Influence of mask annotation location.}
Notably,
in our framework, the annotated mask can be any frame of the video. Tab.\ref{al}
%\scc{Table 1}
%scc:加引用
% investigates the effect of the annotation mask 
% being located at three different locations in the video on the inpainting results. 
investigates the effect of the location of the known annotation mask, where we explore three different locations in the video on the inpainting results.
%scc:加引用,and the tabel should be consistent. Table or Tab
% As shown in Tab.1, the effect of the masks in three positions on the inpainting results is minimal.
As shown in Tab.\ref{al}, the inpainting result difference caused by the location of the mask
% in three positions 
% on the inpainting results 
is minimal and negligible.
%\scc{,,,}.%scc:可以忽略不计的
This shows that our semi-supervised framework is robust to the annotated locations of masks.

\noindent\textbf{Effectiveness of Feature Alignment module.}
In Tab.\ref{AS_FAN_1}, we verify the effectiveness of the feature alignment module constructed with different numbers of DCA blocks. We can see that the feature alignment module constructed using DCA blocks can improve the effect of video inpainting, and the more the number of stacked DCA blocks, the better the performance. Considering the time cost, we stack four layers of DCA blocks in the feature alignment module.

\noindent\textbf{Refinement of Inpainting Results.}
We investigate the impact of annotating different numbers of video frames in the video on inpainting results. As shown in Tab.\ref{Tab4}, the inpainting quality of our method is further improved with the increases of annotated frames. This means that our method can improve the video inpainting quality by increasing the number of annotated frames in some specific environments. 

More ablation experiments can be found in the supplementary materials.

\begin{table}[tb]
\renewcommand\tabcolsep{2.2pt}
\scriptsize
\centering
\caption{Effectiveness of the feature alignment module constructed with different numbers of DCA blocks.}
\begin{tabular}{cccccccc}
\hline %\diagbox
{{N}}  &{0}   &{1}   &{2}   &{3}   &{4}  &{5}   &{6}\\
\hline
 {{PSNR$\uparrow$}}   &{26.667}      &{29.569}     &{30.174}    &{30.525}     &{30.834}      &{30.839}      &{30.851} \\
 {{SSIM$\uparrow$}}   &{0.8580}       &{0.9036}      &{0.9081}    &{0.9124}    &{0.9135}      &{0.9137}      &{0.9156}\\
 {{$E_{warp}\downarrow$}}   &{0.1668}      &{0.1554}     &{0.1511}    &{0.1464} &{0.1452}     &{0.1453}      &{0.1455} \\
 {{LPIPS$\downarrow$}}   &{0.6916}       &{0.4302}      &{0.4286}    &{0.4213}   &{0.4171}     &{0.4173}      &{0.4172}\\
\hline
\end{tabular}
\label{AS_FAN_1}
\end{table}

\begin{table}[!t]
\scriptsize
\centering
\caption{The impact of different numbers of annotated frames in the video on inpainting results.}
\begin{tabular}{ccccccc}
\hline %\diagbox
{{Annotation}}  &{1}   &{2}   &{3}     &{5}   &{7}\\
\hline
PSNR$\uparrow$ &30.834         &30.918      &30.974      &31.086     &31.056   \\
SSIM$\uparrow$  &0.9135          &0.9142       &0.9147   &0.9156      &0.9196  \\
{{$E_{warp}\downarrow$}}   &{0.1452}      &{0.1443}     &{0.1437}    &{0.1430} &{0.1424} \\
 {{LPIPS$\downarrow$}}   &{0.4171}       &{0.4147}      &{0.4124}    &{0.4108}     &{0.4102} \\
\hline
\end{tabular}
\label{Tab4}
\end{table}
\section{Conclusion}
%In this paper, we formulate a new task termed semi-supervised video inpainting and proposed an efficient method to tackle it. Semi-supervised video inpainting is essential for real-world applications since it can significantly reduce annotation costs. Experimental results show that the proposed method is feasible and effective in video inpainting. Notably, our model provides a benchmark for subsequent research on semi-supervised video inpainting.
In this paper, we formulate a new task termed semi-supervised video inpainting and propose an end-to-end trainable framework consisting of 
% which includes a pair of dual networks: 
completion network and mask prediction network to tackle it.
%\scc{an efficient method to tackle it}.
%scc0812:简单说一下是怎样的一个方法
This task is essential for real-world applications since it can significantly reduce annotation 
% costs. 
cost. Experimental results show that the proposed method is effective in semi-supervised video inpainting tasks. Notably, we also tailor a new dataset for the semi-supervised video inpainting task, which can effectively facilitate subsequent research.

{\small
\bibliographystyle{ieee_fullname}
\bibliography{reference}

\begin{thebibliography}{10}\itemsep=-1pt

\bibitem{chang2019free}
Ya-Liang Chang, Zhe~Yu Liu, Kuan-Ying Lee, and Winston Hsu.
\newblock Free-form video inpainting with 3d gated convolution and temporal
  patchgan.
\newblock In {\em Proceedings of the IEEE International Conference on Computer
  Vision (ICCV)}, pages 9066--9075, 2019.

\bibitem{chen2020state}
Xi Chen, Zuoxin Li, Ye Yuan, Gang Yu, Jianxin Shen, and Donglian Qi.
\newblock State-aware tracker for real-time video object segmentation.
\newblock In {\em Proceedings of the IEEE Conference on Computer Vision and
  Pattern Recognition (CVPR)}, pages 9384--9393, 2020.

\bibitem{8237351}
Jifeng Dai, Haozhi Qi, Yuwen Xiong, Yi Li, Guodong Zhang, Han Hu, and Yichen
  Wei.
\newblock Deformable convolutional networks.
\newblock In {\em Proceedings of the IEEE International Conference on Computer
  Vision (ICCV)}, pages 764--773, 2017.

\bibitem{Gao-ECCV-FGVC}
Chen Gao, Ayush Saraf, Jia-Bin Huang, and Johannes Kopf.
\newblock Flow-edge guided video completion.
\newblock In {\em Proceedings of the European Conference on Computer Vision
  (ECCV)}, pages 713--729, 2020.

\bibitem{granados2012how}
Miguel Granados, James Tompkin, Kwang~In Kim, Oliver Grau, Jan Kautz, and
  Christian Theobalt.
\newblock How not to be seen-object removal from videos of crowded scenes.
\newblock {\em Computer Graphics Forum}, 31(2):219--228, 2012.

\bibitem{9008384}
Zhang Haotian, Mai Long, Wang Hailin, JinZha ando~wen, and Ning Xu;~John
  Collomosse.
\newblock An internal learning approach to video inpainting.
\newblock In {\em Proceedings of the IEEE International Conference on Computer
  Vision (ICCV)}, pages 2720--2729, 2019.

\bibitem{8578250}
P. Hu, G. Wang, X. Kong, J. Kuen, and Y. Tan.
\newblock Motion-guided cascaded refinement network for video object
  segmentation.
\newblock In {\em Proceedings of the IEEE Conference on Computer Vision and
  Pattern Recognition (CVPR)}, pages 1400--1409, 2018.

\bibitem{huang2016temporally}
Jiabin Huang, Sing~Bing Kang, Narendra Ahuja, and Johannes Kopf.
\newblock Temporally coherent completion of dynamic video.
\newblock {\em ACM Transactions on Grapics (TOG)}, 35(6):196.1--196.11, 2016.

\bibitem{huang2020fast}
Xuhua Huang, Jiarui Xu, Yu-Wing Tai, and Chi-Keung Tang.
\newblock Fast video object segmentation with temporal aggregation network and
  dynamic template matching.
\newblock In {\em Proceedings of the IEEE Conference on Computer Vision and
  Pattern Recognition (CVPR)}, pages 8879--8889, 2020.

\bibitem{Kang2022ErrorCF}
Jaeyeon Kang, Seoung~Wug Oh, and Seon~Joo Kim.
\newblock Error compensation framework for flow-guided video inpainting.
\newblock In {\em European Conference on Computer Vision}, pages 375--390,
  2022.

\bibitem{Kim_2019_CVPR}
Dahun Kim, Sanghyun Woo, Joon-Young Lee, and In~So Kweon.
\newblock Deep blind video decaptioning by temporal aggregation and recurrence.
\newblock In {\em Proceedings of the IEEE Conference on Computer Vision and
  Pattern Recognition (CVPR)}, pages 4263--4272, 2019.

\bibitem{8931251}
Dahun Kim, Sanghyun Woo, Joon-Young Lee, and In~So Kweon.
\newblock Recurrent temporal aggregation framework for deep video inpainting.
\newblock {\em IEEE Transactions on Pattern Analysis and Machine Intelligence
  (TPAMI)}, 42(5):1038--1052, 2020.

\bibitem{Kim_2022_CVPR}
Soo~Ye Kim, Kfir Aberman, Nori Kanazawa, Rahul Garg, Neal Wadhwa, Huiwen Chang,
  Nikhil Karnad, Munchurl Kim, and Orly Liba.
\newblock Zoom-to-inpaint: Image inpainting with high-frequency details.
\newblock In {\em Proceedings of the IEEE/CVF Conference on Computer Vision and
  Pattern Recognition (CVPR) Workshops}, pages 477--487, 2022.

\bibitem{lai2018learning}
Wei-Sheng Lai, Jia-Bin Huang, Oliver Wang, Eli Shechtman, Ersin Yumer, and
  Ming-Hsuan Yang.
\newblock Learning blind video temporal consistency.
\newblock In {\em Proceedings of the European conference on computer vision
  (ECCV)}, pages 179--195, 2018.

\bibitem{lao2021flow}
Dong Lao, Peihao Zhu, Peter Wonka, and Ganesh Sundaramoorthi.
\newblock Flow-guided video inpainting with scene templates.
\newblock In {\em Proceedings of the IEEE International Conference on Computer
  Vision (ICCV)}, pages 14599--14608, 2021.

\bibitem{lee2019cpnet}
Sungho Lee, Seoung~Wug Oh, DaeYeun Won, and Seon~Joo Kim.
\newblock Copy-and-paste networks for deep video inpainting.
\newblock In {\em Proceedings of the IEEE International Conference on Computer
  Vision (ICCV)}, pages 4413--4421, 2019.

\bibitem{Li2020ShortTermAL}
Ang Li, Shanshan Zhao, Xingjun Ma, M. Gong, Jianzhong Qi, Rui Zhang, Dacheng
  Tao, and R. Kotagiri.
\newblock Short-term and long-term context aggregation network for video
  inpainting.
\newblock In {\em Proceedings of the European Conference on Computer Vision
  (ECCV)}, 2020.

\bibitem{li2022towards}
Zhen Li, Cheng-Ze Lu, Jianhua Qin, Chun-Le Guo, and Ming-Ming Cheng.
\newblock Towards an end-to-end framework for flow-guided video inpainting.
\newblock In {\em Proceedings of the IEEE/CVF Conference on Computer Vision and
  Pattern Recognition (CVPR)}, pages 17562--17571, 2022.

\bibitem{liao2020dvi}
Miao Liao, Feixiang Lu, Dingfu Zhou, Sibo Zhang, Wei Li, and Ruigang Yang.
\newblock Dvi: Depth guided video inpainting for autonomous driving.
\newblock In {\em Proceedings of the European Conference on Computer Vision
  (ECCV)}, pages 1--17, 2020.

\bibitem{liu2021fuseformer}
Rui Liu, Hanming Deng, Yangyi Huang, Xiaoyu Shi, Lewei Lu, Wenxiu Sun, Xiaogang
  Wang, Jifeng Dai, and Hongsheng Li.
\newblock Fuseformer: Fusing fine-grained information in transformers for video
  inpainting.
\newblock In {\em Proceedings of the IEEE International Conference on Computer
  Vision (ICCV)}, pages 14040--14049, 2021.

\bibitem{9558783}
Ruixin Liu, Bairong Li, and Yuesheng Zhu.
\newblock Temporal group fusion network for deep video inpainting.
\newblock {\em IEEE Transactions on Circuits and Systems for Video Technology},
  32(6):3539--3551, 2022.

\bibitem{liu2020temporal}
Ruixin Liu, Zhenyu Weng, Yuesheng Zhu, and Bairong Li.
\newblock Temporal adaptive alignment network for deep video inpainting.
\newblock In {\em International Joint Conference on Artificial Intelligence
  (IJCAI)}, pages 927--933, 2020.

\bibitem{lu2020video}
Xiankai Lu, Wenguan Wang, Danelljan Martin, Tianfei Zhou, Jianbing Shen, and
  Van~Gool Luc.
\newblock Video object segmentation with episodic graph memory networks.
\newblock In {\em Proceedings of the European Conference on Computer Vision
  (ECCV)}, pages 661--679, 2020.

\bibitem{newson2014video}
Alasdair Newson, Andres Almansa, Matthieu Fradet, Yann Gousseau, and Patrick
  Perez.
\newblock Video inpainting of complex scenes.
\newblock {\em SIAM Journal on Imaging Sciences}, 7(4):1993--2019, 2014.

\bibitem{7780454}
F. Perazzi, J. Pont-Tuset, B. McWilliams, L.~Van Gool, M. Gross, and A.
  Sorkine-Hornung.
\newblock A benchmark dataset and evaluation methodology for video object
  segmentation.
\newblock In {\em Proceedings of the IEEE Conference on Computer Vision and
  Pattern Recognition (CVPR)}, pages 724--732, 2016.

\bibitem{Perazzi2016A}
F. Perazzi, J. Pont-Tuset, B. Mcwilliams, L.~Van Gool, and A. Sorkine-Hornung.
\newblock A benchmark dataset and evaluation methodology for video object
  segmentation.
\newblock In {\em Proceedings of the IEEE Conference on Computer Vision and
  Pattern Recognition (CVPR)}, pages 724--732, 2016.

\bibitem{Ren_2022_CVPR}
Jingjing Ren, Qingqing Zheng, Yuanyuan Zhao, Xuemiao Xu, and Chen Li.
\newblock Dlformer: Discrete latent transformer for video inpainting.
\newblock In {\em Proceedings of the IEEE/CVF Conference on Computer Vision and
  Pattern Recognition (CVPR)}, pages 3511--3520, 2022.

\bibitem{robinson2020learning}
Andreas Robinson, Felix~Jaremo Lawin, Martin Danelljan, Fahad~Shahbaz Khan, and
  Michael Felsberg.
\newblock Learning fast and robust target models for video object segmentation.
\newblock In {\em Proceedings of the IEEE Conference on Computer Vision and
  Pattern Recognition (CVPR)}, pages 7406--7415, 2020.

\bibitem{seong2021hierarchical}
Hongje Seong, Seoung~Wug Oh, Joon-Young Lee, Seongwon Lee, Suhyeon Lee, and
  Euntai Kim.
\newblock Hierarchical memory matching network for video object segmentation.
\newblock In {\em Proceedings of the IEEE International Conference on Computer
  Vision (ICCV)}, pages 12889--12898, 2021.

\bibitem{9008790}
Oh Seoung, Wug, Lee Joon-Young, Xu Ning, and Kim Seon, Joo.
\newblock Video object segmentation using space-time memory networks.
\newblock In {\em Proceedings of the IEEE International Conference on Computer
  Vision (ICCV)}, pages 9225--9234, 2019.

\bibitem{9010390}
Oh Seoung, Wug, Lee Sungho, Lee Joon-Young, and Kim Seon, Joo.
\newblock Onion-peel networks for deep video completion.
\newblock In {\em Proceedings of the IEEE International Conference on Computer
  Vision (ICCV)}, pages 4402--4411, 2019.

\bibitem{sun2020fast}
Mingjie Sun, Jimin Xiao, Eng~Gee Lim, Bingfeng Zhang, and Yao Zhao.
\newblock Fast template matching and update for video object tracking and
  segmentation.
\newblock In {\em Proceedings of the IEEE Conference on Computer Vision and
  Pattern Recognition (CVPR)}, pages 10791--10799, 2020.

\bibitem{wang2018video}
Chuan Wang, Haibin Huang, Xiaoguang Han, and Jue Wang.
\newblock Video inpainting by jointly learning temporal structure and spatial
  details.
\newblock In {\em Proceedings of the AAAI Conference on Artificial Intellignce
  (AAAI)}, pages 5232--5239, 2019.

\bibitem{wang2019fast}
Qiang Wang, Li Zhang, Luca Bertinetto, Weiming Hu, and Philip~HS Torr.
\newblock Fast online object tracking and segmentation: A unifying approach.
\newblock In {\em Proceedings of the IEEE conference on Computer Vision and
  Pattern Recognition (CVPR)}, pages 1328--1338, 2019.

\bibitem{978-3-030-58595-2_45}
Yi Wang, Ying-Cong Chen, Xin Tao, and Jiaya Jia.
\newblock Vcnet: A robust approach to blind image inpainting.
\newblock In Andrea Vedaldi, Horst Bischof, Thomas Brox, and Jan-Michael Frahm,
  editors, {\em Proceedings of the European Conference on Computer Vision
  (ECCV)}, pages 752--768, 2020.

\bibitem{wang2018image}
Yi Wang, Xin Tao, Xiaojuan Qi, Xiaoyong Shen, and Jiaya Jia.
\newblock Image inpainting via generative multi-column convolutional neural
  networks.
\newblock {\em Advances in neural information processing systems (NIPS)}, 31,
  2018.

\bibitem{Wang_2019_ICCV}
Ziqin Wang, Jun Xu, Li Liu, Fan Zhu, and Ling Shao.
\newblock Ranet: Ranking attention network for fast video object segmentation.
\newblock In {\em Proceedings of the IEEE International Conference on Computer
  Vision (ICCV)}, October 2019.

\bibitem{9446636}
Zhiliang Wu, Kang Zhang, Hanyu Xuan, Jian Yang, and Yan Yan.
\newblock Dapc-net: Deformable alignment and pyramid context completion
  networks for video inpainting.
\newblock {\em IEEE Signal Processing Letters}, 28:1145--1149, 2021.

\bibitem{xie2021efficient}
Haozhe Xie, Hongxun Yao, Shangchen Zhou, Shengping Zhang, and Wenxiu Sun.
\newblock Efficient regional memory network for video object segmentation.
\newblock In {\em Proceedings of the IEEE Conference on Computer Vision and
  Pattern Recognition (CVPR)}, pages 1286--1295, 2021.

\bibitem{Xu2018YouTube}
Ning Xu, Linjie Yang, Yuchen Fan, Jianchao Yang, Dingcheng Yue, Yuchen Liang,
  Brian Price, Scott Cohen, and Thomas Huang.
\newblock Youtube-vos: Sequence-to-sequence video object segmentation.
\newblock In {\em Proceedings of the European Conference on Computer Vision
  (ECCV)}, pages 603--619, 2018.

\bibitem{xu2019deep}
Rui Xu, Xiaoxiao Li, Bolei Zhou, and Chen~Change Loy.
\newblock Deep flow-guided video inpainting.
\newblock In {\em Proceedings of the IEEE Conference on Computer Vision and
  Pattern Recognition (CVPR)}, pages 3723--3732, 2019.

\bibitem{9710199}
Yingchen Yu, Fangneng Zhan, Shijian Lu, Jianxiong Pan, Feiying Ma, Xuansong
  Xie, and Chunyan Miao.
\newblock Wavefill: A wavelet-based generation network for image inpainting.
\newblock In {\em IEEE International Conference on Computer Vision (ICCV)},
  pages 14094--14103, 2021.

\bibitem{9008825}
X. Zeng, R. Liao, L. Gu, Y. Xiong, S. Fidler, and R. Urtasun.
\newblock Dmm-net: Differentiable mask-matching network for video object
  segmentation.
\newblock In {\em Proceedings of the IEEE International Conference on Computer
  Vision (ICCV)}, pages 3928--3937, 2019.

\bibitem{yan2020sttn}
Yanhong Zeng, Jianlong Fu, and Hongyang Chao.
\newblock Learning joint spatial-temporal transformations for video inpainting.
\newblock In {\em Proceedings of the European Conference on Computer Vision
  (ECCV)}, pages 3723--3732, 2020.

\bibitem{yan2019PENnet}
Yanhong Zeng, Jianlong Fu, Hongyang Chao, and Baining Guo.
\newblock Learning pyramid-context encoder network for high-quality image
  inpainting.
\newblock In {\em Proceedings of the IEEE Conference on Computer Vision and
  Pattern Recognition (CVPR)}, pages 1486--1494, 2019.

\bibitem{zhang2022flow}
Kaidong Zhang, Jingjing Fu, and Dong Liu.
\newblock Flow-guided transformer for video inpainting.
\newblock {\em Proceedings of the European Conference on Computer Vision
  (ECCV)}, 2022.

\bibitem{Zhang_2022_CVPR}
Kaidong Zhang, Jingjing Fu, and Dong Liu.
\newblock Inertia-guided flow completion and style fusion for video inpainting.
\newblock In {\em Proceedings of the IEEE/CVF Conference on Computer Vision and
  Pattern Recognition (CVPR)}, pages 5982--5991, 2022.

\bibitem{zhang2019fast}
Lu Zhang, Zhe Lin, Jianming Zhang, Huchuan Lu, and You He.
\newblock Fast video object segmentation via dynamic targeting network.
\newblock In {\em Proceedings of the IEEE International Conference on Computer
  Vision (ICCV)}, pages 5582--5591, 2019.

\bibitem{zhang2018unreasonable}
Richard Zhang, Phillip Isola, Alexei~A Efros, Eli Shechtman, and Oliver Wang.
\newblock The unreasonable effectiveness of deep features as a perceptual
  metric.
\newblock In {\em Proceedings of the IEEE Conference on Computer Vision and
  Pattern Recognition (CVPR)}, pages 586--595, 2018.

\bibitem{zou2021progressive}
Xueyan Zou, Linjie Yang, Ding Liu, and Yong~Jae Lee.
\newblock Progressive temporal feature alignment network for video inpainting.
\newblock In {\em Proceedings of the IEEE Conference on Computer Vision and
  Pattern Recognition (CVPR)}, pages 16448--16457, 2021.

\end{thebibliography}
}

\end{document}